\documentclass[letterpaper, 10 pt, conference]{ieeeconf}  

\IEEEoverridecommandlockouts                              

\usepackage[
  colorlinks=true,
  linkcolor=magenta,
  urlcolor=magenta
]{hyperref}
\usepackage{amsmath,amsfonts}
\usepackage{float}
\usepackage{algorithm}
\usepackage{algorithmic}

\usepackage{array}
\usepackage[caption=false,font=normalsize,labelfont=sf,textfont=sf]{subfig}
\usepackage{textcomp}
\usepackage{stfloats}
\usepackage{url}
\usepackage{verbatim}
\usepackage{graphicx}
\usepackage{cite}
\usepackage{multirow}
\usepackage{booktabs}
\usepackage{ifthen}
\usepackage{siunitx}
\usepackage{amssymb}
\usepackage{makecell}

\usepackage{graphicx}
\usepackage{lineno}
\usepackage{soulutf8} 
\usepackage{tcolorbox}

\title{\LARGE \bf
Multimodal Adversarial Quality Policy for Safe Grasping
}

\author{Kunlin Xie$^*$,  Chenghao Li$^*$, Haolan Zhang, and Nak Young Chong
\thanks{The authors are with the School of Information Science, Japan Advanced Institute of Science and Technology, Nomi, Ishikawa 923-1292, Japan {\tt\small \{s2410074, chenghao.li, haolan.z, nakyoung\}@jaist.ac.jp}.}
\thanks{$^*$These authors contributed equally to this work.}
}

\begin{document}

\maketitle
\thispagestyle{empty}
\pagestyle{empty}

\begin{abstract}

Vision-guided robot grasping based on Deep Neural Networks (DNNs) generalizes well but poses safety risks in the Human-Robot Interaction (HRI). Recent works solved it by designing benign adversarial attacks and patches with RGB modality, yet depth-independent characteristics limit their effectiveness on RGBD modality. In this work, we propose the Multimodal Adversarial Quality Policy (MAQP) to realize multimodal safe grasping. Our framework introduces two key components. First, the Heterogeneous Dual-Patch Optimization Scheme (HDPOS) mitigates the distribution discrepancy between RGB and depth modalities in patch generation by adopting modality-specific initialization strategies, employing a Gaussian distribution for depth patches and a uniform distribution for RGB patches, while jointly optimizing both modalities under a unified objective function. Second, the Gradient-Level Modality Balancing Strategy (GLMBS) is designed to resolve the optimization imbalance from RGB and Depth patches in patch shape adaptation by reweighting gradient contributions based on per-channel sensitivity analysis and applying distance-adaptive perturbation bounds. We conduct extensive experiments on the benchmark datasets and a cobot, showing the effectiveness of MAQP.
\end{abstract}

\section{INTRODUCTION}
\begin{figure}[!t]
\vspace{0.5\baselineskip}
\centerline{\includegraphics[width=\columnwidth]{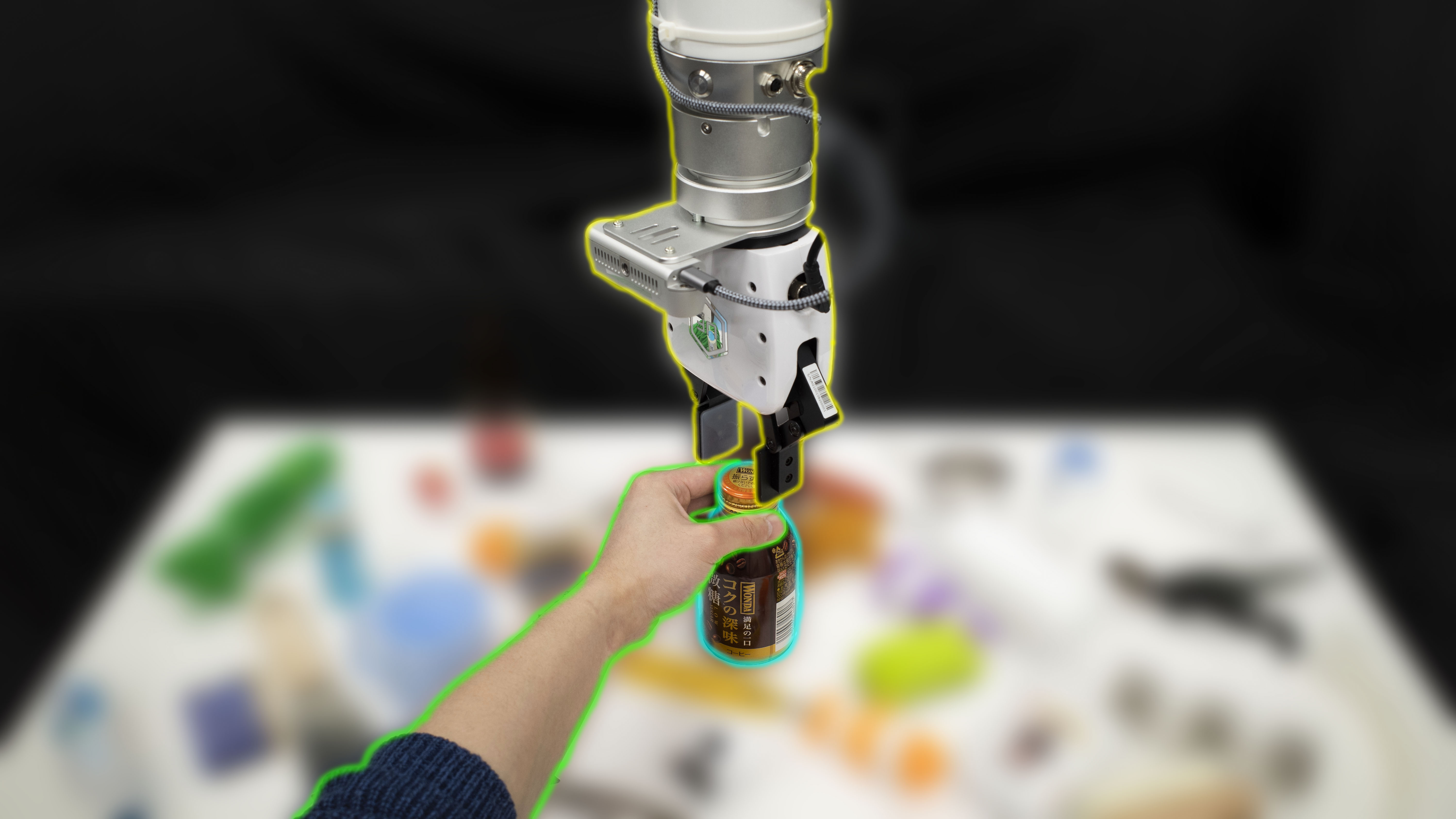}}
\caption{An example of safety risk in HRI: the robot mistakenly identifies the human hand or adjacent objects as graspable targets for grasping, causing potential injury to the human worker.}
\label{fig1}
\end{figure}

Visual grasping is a core capability for robotics in HRI scenarios \cite{c1}, aimed at helping humans improve work efficiency in the service and manufacturing domains. Traditional approaches \cite{c1, c2, c3} often rely on explicit object models and analytic metrics, which are limited in unstructured scenes with unknown objects. Recent DNNs-based methods \cite{c4, c5, c6, c7, c8} predict grasps directly from images and have shown strong generalization to unknown objects. However, this generalizability can introduce safety risks, that is grasping model may assign high grasp confidence to human hands or nearby objects in HRI, as shown in Fig. \ref{fig1}.

To mitigate this safety risk, Li {\it et al.} \cite{c9} designed the Quality-focused Active Adversarial Policy (QFAAP) from the benign adversarial attack perspective, which proposes the Adversarial Quality Patch (AQP) together with the Projected Quality Gradient Descent (PQGD) to manipulate grasp quality scores, thereby steering grasping away from unsafe regions. Specifically, the AQP is used to generate adversarial patches by the grasping model and dataset, and the PQGD is used to refine the patches and endow them with shape adaptation for human hands. However, QFAAP was validated only for RGB-based grasping models, while visual grasping platforms often rely on RGBD sensing. More importantly, realizing grasp quality manipulation to RGBD is non-trivial because of the distribution discrepancy and optimization imbalance between depth and RGB information.

Therefore, in this work, we present the Multi-modal Adversarial Quality Policy (MAQP) framework, which is based on a comprehensive understanding of the distinctive characteristics of RGB and depth modalities. To address the distribution discrepancy between modalities in the patch generation process, we adopt modality-specific initialization strategies for RGB and depth while jointly training an RGBD adversarial quality patch under a unified objective function. To mitigate the optimization imbalance across modalities in the patch shape adaptation process, we reweight gradient contributions according to per-channel sensitivity analysis and apply distance-adaptive perturbation bounds that incorporate the physical noise characteristics of the depth sensor. The contributions of this work are as follows:

\begin{itemize}
    \item We propose the HDPOS by modality-specific initialization. Gaussian initialization aligned with zero-centered preprocessing for depth, and uniform initialization with non-negative normalization and heterogeneous constraints for RGB, while jointly optimizing both modalities under a unified objective function.
    
    \item We introduce the GLMBS by a gradient reweighting mechanism. Based on the analysis of depth sensitivity, GLMBS inversely reweights gradients according to modality sensitivity and incorporates distance-adaptive perturbation bounds reflecting depth sensor noise.
    
    \item The results demonstrate our modality-aware design is critical for effectively manipulating grasp quality scores in RGBD-based robotic grasping systems and have strong potential for generalization to broader multimodal robotic manipulation tasks.
\end{itemize}

This paper is structured as follows. Section \ref{sec:related} (Related Work) provides a review of RGBD perception for robot grasping and adversarial attacks. Section \ref{sec:qfaap} (Quality-focused Active Adversarial Policy) describes the concept of QFAAP. Section \ref{sec:method} (Proposed Method) presents an overview of MAQP and elaborates on its two core components, HDPOS and GLMBS, together with their respective submodules. Section \ref{sec:ex} (Experiments) evaluates the proposed approach across different grasping models and datasets, conducts ablation studies to assess the contribution of each component, and further validates the method in real-world robot grasping scenarios. Finally, Section \ref{sec:con} (Conclusion) summarizes the main findings and outlines directions for future research.

\section{Related Work} \label{sec:related}
\subsection{DNNs-based Visual Grasping}
Although numerous robot grasping frameworks have been proposed, this work specifically investigates DNNs-based 4-Degree-of-Freedom (4-DOF) visual grasping with a parallel-jaw gripper. These approaches exhibit strong generalization to previously unseen objects by training deep function approximators on large-scale datasets to directly infer grasp success probabilities from visual observations. In addition, RGBD perception is widely adopted in such methods because of its complementary properties: RGB images convey rich appearance and texture information, whereas depth data provide explicit geometric structure.

Along this line of research, \cite{c4} presented a multimodal deep CNN that maps RGBD image pairs of novel objects to optimal grasping. Kumra {\it et al.} \cite{c5} employed a ResNet-based architecture \cite{c6} to process RGBD inputs for grasp detection. From a region-proposal perspective, Chu {\it et al.} \cite{c7} incorporated a Region Proposal Network (RPN) \cite{c8} to enable simultaneous prediction of multiple grasps across multiple objects using RGBD data. Asif {\it et al.} \cite{c10} proposed EnsembleNet, which produces four grasp representations and aggregates them to compute grasp quality scores, selecting the candidate with the highest score. Yan {\it et al.} \cite{c11} adopted a point-cloud-based CNN for grasp generation, where color, depth, and masked images are first processed to reconstruct a 3D point cloud of the target object, which is then fed into a dedicated prediction network. Moreover, Kumra {\it et al.} \cite{c12} introduced a generative residual convolutional neural network capable of real-time synthesis of robust antipodal grasps from $n$-channel inputs.

Despite demonstrating impressive generalization in unstructured environments, these methods emphasize grasp generalizability while overlooking safety considerations. Their object-agnostic nature may lead them to misidentify human hands or nearby objects as valid grasp targets, thereby introducing potential safety risks in HRI scenarios.

\subsection{Adversarial Attacks}
Since Szegedy {\it et al.} \cite{c13} first revealed the existence of adversarial examples, substantial efforts have been made to investigate the vulnerability of DNNs. Existing studies can be broadly classified into two categories: single-image adversarial attacks and image-agnostic attacks, also known as adversarial patch attacks. Single-image adversarial attacks generate global perturbations over the entire image by maximizing the model’s discriminative loss. Goodfellow {\it et al.} \cite{c14} introduced the Fast Gradient Sign Method (FGSM), which leverages the approximate linearity of neural networks to produce effective perturbations in a single step. Building upon this idea, Wang {\it et al.} \cite{c15} and Madry {\it et al.} \cite{c16} reformulated the one-step perturbation of FGSM into an iterative optimization process, resulting in I-FGSM and Projected Gradient Descent (PGD), respectively. While these single-image attacks can efficiently mislead image classification models, they are inherently input-specific and typically modify the entire image, requiring re-optimization for each new sample. Such limitations motivate the development of more versatile approaches capable of targeting arbitrary images and localized regions.

Adversarial patch attacks, in contrast, emphasize spatial locality and image-agnostic characteristics, making them particularly effective against object detection models with localization functionality. For instance, Liu {\it et al.} \cite{c17} proposed DPatch to attack widely used object detectors, reducing detection accuracy and inducing mislocalization or misclassification. Lee {\it et al.} \cite{c18} further examined the weaknesses of DPatch and proposed Robust DPatch to enhance attack stability. Beyond degrading detection performance, subsequent studies \cite{c19, c20} explored detection evasion, where adversarial patches prevent detectors from identifying occluded objects. Later works \cite{c21, c22, c23} extended adversarial patches into adversarial clothing by replicating the patterns onto garments, thereby enabling more robust attacks under varying viewpoints. However, such replication-based extensions incur high costs and are generally limited by the deformation characteristics of clothing materials.

More recently, Li {\it et al.} \cite{c9} integrated the strengths of single-image adversarial and adversarial patch attacks to propose an active adversarial framework with rapid adaptability to diverse human hand shapes. By directly manipulating grasp quality scores, their method addresses safety concerns in DNN-based grasping systems during HRI. Nevertheless, their validation was confined to RGB-based grasping models, whereas most visual grasping systems typically rely on RGBD sensing. More importantly, achieving grasp quality manipulation by benign adversarial attacks in the RGBD modality is challenging due to the inherent distribution discrepancies and optimization imbalance between the RGB and depth modalities.

\section{QFAAP} \label{sec:qfaap}
In this section, we recall the QFAAP \cite{c9}, which is based on the RGB modality. Specifically, two main modules of QFAAP, the AQP and PQGD, will be explained here.

\subsection{Adversarial Quality Patch}

 In the generation process of AQP ($\mathbf{p}_{t}$), the training RGB images were normalized to the range [0, 1], and the AQP was initialized from a uniform distribution, as shown in Eq. \ref{eq:1}, where $H \times W$ represents the size of the patch. $\mathcal{U}(0, 1)$ denotes the continuous uniform distribution over the interval [0, 1].

\begin{equation}
\label{eq:1}
\mathbf{p}_{t} \sim \mathcal{U}(0, 1)^{3 \times H \times W}
\end{equation}

As the training process updated the AQP, the optimization could drive the adversarial patch outside the valid range. To satisfy the physical realizability constraint, the clipping operation was applied after each update to project the AQP back into the valid range, as shown in Eq. \ref{eq:2}. Here, $\Pi$ denotes the projection operator.

\begin{equation}
\label{eq:2}
\mathbf{p}_{t} = \Pi_{[0,1]}(\mathbf{p}_{t})
\end{equation}

The objective function consisted of three components. The quality loss $\mathcal{L}_q^p$ is designed to stably maximize the mean quality score within the patch region, as shown in Eq. \ref{eq:3}, where the quality score within the AQP area is denoted as $\mathcal{Q}_i^p$, and $\alpha$ controll the trade-off between maximizing the mean quality $\mathbb{E}(\mathcal{Q}_i^p)$ and minimizing the variance $\text{Var}(\mathcal{Q}_i^p)$. The negative sign converts the maximization objective into a minimization problem compatible with gradient descent.

\begin{equation}
\label{eq:3}
\mathcal{L}_q^p = \frac{1}{B}\sum\limits_{i=1}^{B}[-\mathbb{E}(\mathcal{Q}_i^p)+\alpha\text{Var}(\mathcal{Q}_i^p)]
\end{equation}

The difference loss $\mathcal{L}_{d}$ encourages the quality within the patch region to exceed that of the surrounding areas, as shown in Eq. \ref{eq:4}, where the quality score outside the AQP area is denoted as $\mathcal{\tilde{Q}}_i^p$. This formulation ensures that even the worst-performing pixel within the patch achieves a quality comparable to the best pixel outside it.

\begin{equation}
\label{eq:4}
\mathcal{L}_{d} = \frac{1}{B} \sum_{i=1}^{B} \left|\min\mathcal{Q}_i^p-\max\mathcal{\tilde{Q}}_i^p\right|
\end{equation}

The total variation is adopted to mitigate the noise introduced during AQP training, consistent with \cite{c20}, as shown in Eq. \ref{eq:5}. Here, $\mathbf{p}_t(j^p, k^p)$ represent the pixel value of AQP at location $(j^p, k^p)$.

\begin{equation}
\label{eq:5}
\mathcal{L}_{tv} = \frac{1}{H \times W} \sum_{j^p=1}^{H} \sum_{k^p=1}^{W} \left\| \mathbf{p}_t(j^p, k^p) \right\|_2
\end{equation}

The complete loss function combines these components, as shown in Eq. \ref{eq:6}, where $\beta$ and $\gamma$ control the relative importance of the difference loss and the total variation loss, respectively.

\begin{equation}
\label{eq:6}
\mathcal{L}_{aqp} = \mathcal{L}_q^p + \beta\mathcal{L}_{tv} + \gamma\mathcal{L}_{d}
\end{equation}

\subsection{Projected Quality Gradient Descent}

The PQGD had been employed to rapidly refine the AQP within the masked hand region during the shape adaptation process. First, $\mathbf{x'}$ is defined as the RGB frame after inserting the AQP into the hand area, as shown in Eq. \ref{eq:7}. Here, $\mathbf{x}$ denotes an RGB frame captured by a depth camera, and $\mathcal{M}_h$ represents the hand mask associated with $\mathbf{x}$.

\begin{equation}
\label{eq:7}
\mathbf{x}' = \mathbf{x}(1-\mathcal{M}_h) + \mathbf{p}_t\mathcal{M}_h
\end{equation}

Then, the PQGD loss $\mathcal{L}_{pqgd}$ is defined as shown in Eq. \ref{eq:8}, where $\mathbf{Q}_t^h$ represents the quality map inside the hand region of $\mathbf{x}_t''$, and $\mathbf{x}_t''$ denotes the RGB frame after incorporating both AQP and PQGD within the hand area.

\begin{equation}
\label{eq:8}
\mathcal{L}_{pqgd} = -\mathbb{E}(\mathbf{Q}^h_{t-1})
\end{equation}

Finally, $\mathcal{L}_{pqgd}$ together with the hand mask $\mathcal{M}_h$ were leveraged to refine the AQP within the hand region of $\mathbf{x}_t''$, as shown in Eq. \ref{eq:9}. Here, $\text{sgn}$ denotes the sign function, which computes the direction of the derivative of $\mathcal{L}_{pqgd}$ with respect to $\mathbf{x}''_{t-1}$. The parameter $\delta_{pqgd}$ represents the learning rate.

\begin{equation}
\label{eq:9}
\mathbf{x}_t'' = \left\{\prod_{\mathbf{x}', \epsilon}[\mathbf{x}_{t-1}''- \text{sgn}(\delta_{pqgd}\frac{\partial\mathcal{L}_{pqgd}}{\partial\mathbf{x}_{t-1}''})]\right\}\mathcal{M}_h + \mathbf{x}'(1-\mathcal{M}_h)
\end{equation}

The parameter $\epsilon$ denotes the perturbation bound, which constrains $\mathbf{x}_t''$ from deviating excessively from $\mathbf{x}'$ during refinement, as shown in Eq. \ref{eq:10}.

\begin{equation}
\label{eq:10}
\mathbf{x}_t'' = \Pi_{[{\mathbf{x}'}-\epsilon,, {\mathbf{x}'}+\epsilon]} (\mathbf{x}_t'')
\end{equation}

\section{Proposed Method} \label{sec:method}
In this section, we first provide an overview of the MAQP framework. Subsequently, we present a detailed description of its two key modules, the HDPOS and GLMBS. Finally, we elaborate on the deployment of MAQP for enhancing visual grasping safety in HRI scenarios.

\subsection{Overview of MAQP}
The MAQP framework integrates two tightly coupled components, HDPOS and GLMBS, to achieve effective RGBD grasp quality score manipulation. HDPOS accounts for modality-specific distribution differences by separately initializing RGB and depth patches ($\mathcal{U}(0,1)$ and $\mathcal{N}(0,\sigma_p)$), and using a unified objective function to generate the RGBD patch. GLMBS further addresses optimization imbalance in the patch shape adaptation process by reweighting gradients according to the sensitivity ratio $\rho$, scaling RGB gradients to approach the depth gradients, and introducing a distance-adaptive depth perturbation bound $ \epsilon'(d)$ to further enhance balance. Together, these strategies ensure modality-aware and optimization-balanced benign adversarial learning for RGBD-based safe grasping. We show a pseudocode of MAQP in Algorithm \ref{alg:maqp}.

\subsection{Heterogeneous Dual-Patch Optimization Scheme}
The HDPOS is operated in the patch generation process. We first initialize the patches in each modality by modality-specific initialization strategies to solve the distribution discrepancy between RGB and depth, that is gaussian distribution $\mathcal{N}(0,\sigma_p)$ for depth patches $\mathbf{p}_{d}$ and a uniform distribution $\mathcal{U}(0,1)$ for RGB patches $\mathbf{p}_{rgb}$. The depth patch values are clipped to [0, 1] after sampling, in the same range of RGB patch. This heterogeneous initialization ensures each modality starts from a distribution matching its preprocessed data characteristics. As shown in Eq. \ref{eq:11}:

\begin{equation}
\label{eq:11}
\mathbf{p}_{rgb} \sim \mathcal{U}(0, 1)^{3 \times H \times W}, \quad
\mathbf{p}_{d} \sim \mathcal{N}(0, \sigma_p)^{1 \times H\times W}
\end{equation}

The RGB and depth patches are then applied at the same spatial location of the RGB and depth image pair by a shared mask $\mathcal{M}$, as shown in Eq. \ref{eq:12} and Eq. \ref{eq:13}:

\begin{equation}
\label{eq:12}
    \mathbf{x}_{rgb}' = \mathbf{x}_{rgb} (1 - \mathcal{M}) + \mathbf{p}_{rgb} \mathcal{M}
\end{equation}

\begin{equation}
\label{eq:13}
    \mathbf{x}_d' = \mathbf{x}_d (1 - \mathcal{M}) + \mathbf{p}_d \mathcal{M}
\end{equation}

Therefore, the patched RGBD image $\mathbf{x}_{rgbd}'$ can be formed by channel-wise concatenation:
\begin{equation}
\label{eq:14}
    \mathbf{x}_{rgbd}' = [\mathbf{x}_{rgb}'; \mathbf{x}_d']_c \in \mathbb{R}^{4 \times H \times W}
\end{equation}

\begin{figure}[!t]
\vspace{-0.3\baselineskip}
\centering
\subfloat{\includegraphics[width=0.48\linewidth]{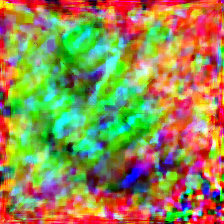}}
\hspace{0.02\linewidth}
\subfloat{\includegraphics[width=0.48\linewidth]{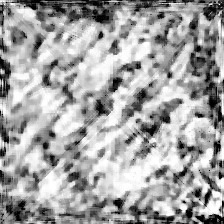}}
\caption{Generated RGBD patch example (RGB and depth patch pair). The left is the RGB patch, the right is the depth patch.}
\label{fig2}
\end{figure}

Finally, these RGBD images $\mathbf{x}_{rgbd}'$ can be used to jointly train both patches through a unified loss function $\mathcal{L}_{aqp}$, the same as Eq. \ref{eq:6}. One generated RGBD patch example is shown in Fig. \ref{fig2}. The unified loss is reasonable because the two patches have the same goal, that is maximize quality scores in the patch region. So the gradients can be fused into a single grasp quality score map by the RGBD grasping model, and backpropagated to both patches according to their contribution. Notably, the constraints for each patch will also be maintained in [0,1] after each iteration of the training. 

\subsection{Gradient-Level Modality Balancing Strategy}
After generating the RGB and depth patch pair by HDPOS, the GLMBS can be used in the patch shape adaptation process to solve the optimization imbalance problems between RGB and depth patches. Specifically, the imbalance issue is induced by the sensitivity bias exacerbated by rapid shape adaptation. Since the visual grasping model is inherently dependent on the geometric information of the objects based on the depth information, grasping models will be more sensitive to depth features than RGB, especially in the shape adaptation process of the patch with only a few iterations. To better exploit the appearance and texture information provided by RGB images for training the RGBD grasping model, it is necessary to balance the gradient contributions of RGB and depth features during this process.

Therefore, we first define per-parameter sensitivity as the average gradient magnitude per channel for RGB $S_{rgb}$ and depth $S_{d}$, as shown in Eq. \ref{eq:15}. Here, $\mathbf{x}'_{rgb}$ and $\mathbf{x}'_{d}$ denote the RGB and depth image after add RGB and depth patches within the human hand area, $\mathcal{L}_{pqgd}$ is the same loss in Eq. \ref{eq:8}, $\nabla$ denotes the gradient calculation.

\begin{equation}
\label{eq:15}
S_{rgb} = \frac{\|\nabla_{\mathbf{x}'_{rgb}} \mathcal{L}_{{pqgd}}\|_2}{3}, 
\quad 
S_d = \|\nabla_{\mathbf{x}'_d} \mathcal{L}_{{pqgd}}\|_2
\end{equation}

Then, the sensitivity ratio $\rho$ is defined in Eq. \ref{eq:16}.

\begin{equation}
\label{eq:16}
    \rho = \frac{S_d}{S_{rgb}}
\end{equation}

\begin{algorithm}[!t]
\caption{Multimodal Adversarial Quality Policy}
\label{alg:maqp}
\begin{algorithmic}[1]
\STATE \textbf{Inputs:} Dataset $\mathcal{D}$, RGBD grasping model $f_\theta$
\STATE \textbf{Outputs:} Optimized patch pair $(\mathbf{p}'_{rgb}, \mathbf{p}'_d)$ in human hand region
\STATE {// HDPOS: $\mathbf{Q}^p$ denote quality score within the patch area, $E$ is epoch number}
\STATE $\mathbf{p}_{rgb} \sim \mathcal{U}(0, 1)^{3 \times H \times W}$
\STATE $\mathbf{p}_d \sim \mathcal{N}(0, \sigma_p)^{1 \times H \times W}$
\FOR{$\text{epoch} = 1$ \TO $E$}
    \FOR{each batch $\{(\mathbf{x}_{rgb}^{(i)}, \mathbf{x}_d^{(i)})\} \in \mathcal{D}$}
        \STATE $\mathbf{x}_{rgb}' \leftarrow \mathbf{x}_{rgb}, \mathbf{p}_{rgb}, \mathcal{M}$
        \STATE $\mathbf{x}_d' \leftarrow \mathbf{x}_d, \mathbf{p}_d, \mathcal{M}$
        \STATE $\mathbf{x}_{rgbd}' \leftarrow [\mathbf{x}_{rgb}'; \mathbf{x}_d']_c$
        \STATE $\mathbf{Q}^p \leftarrow f_\theta(\mathbf{x}_{rgbd}')$
        \STATE $\mathcal{L}_{aqp} \leftarrow \mathbf{Q}^p$
        \STATE $\nabla_{\mathbf{p}_{rgb}} \mathcal{L}_{aqp} \leftarrow \nabla, \mathbf{p}_{rgb}, \mathcal{L}_{aqp}$
        \STATE $\nabla_{\mathbf{p}_{d}} \mathcal{L}_{aqp} \leftarrow \nabla, \mathbf{p}_{d}, \mathcal{L}_{aqp}$
        \STATE $Updated \, \, {\mathbf{p}_{rgb}} \leftarrow \mathbf{p}_{rgb}, \nabla_{\mathbf{p}_{rgb}} \mathcal{L}_{aqp}$
        \STATE $Updated \, \, \mathbf{p}_d \leftarrow \mathbf{p}_d, \nabla_{\mathbf{p}_d} \mathcal{L}_{aqp}$
    \ENDFOR
\ENDFOR
\STATE {// GLMBS: $\mathbf{f}_{rgb}$ and $\mathbf{f}_{d}$ are real-time frame from depth camera, $\mathcal{M}_h$ is the human hand mask, $\mathbf{Q}^h$ is the quality score within human hand area, $I$ is iteration number}
\FOR{$\text{iteration} = 1$ \TO $I$}
        \STATE $\mathbf{f}_{rgb}' \leftarrow \mathbf{p}_{rgb}, \mathbf{f}_{rgb}, \mathcal{M}_h$
        \STATE $\mathbf{f}_{d}' \leftarrow \mathbf{p}_{d}, \mathbf{f}_{d}, \mathcal{M}_h$
        \STATE $\mathbf{f}_{rgbd}' \leftarrow [\mathbf{f}_{rgb}'; \mathbf{f}_d']_c$
        \STATE $\mathbf{Q}^h \leftarrow f_\theta(\mathbf{f}_{rgbd}')$
        \STATE $\mathcal{L}_{pqgd} \leftarrow \mathbf{Q}^h$
        \STATE $\nabla_{\mathbf{p}'_{rgb}} \mathcal{L}_{pqgd} \leftarrow \nabla, \mathbf{p}'_{rgb}, \mathcal{L}_{pqgd}$
        \STATE $\nabla_{\mathbf{p}'_{d}} \mathcal{L}_{pqgd} \leftarrow \nabla, \mathbf{p}'_{d}, \mathcal{L}_{pqgd}$
        \STATE $Balanced \, \, \nabla_{\mathbf{P}'_{rgb}} \mathcal{L}_{pqgd} \leftarrow w_{rgb}, \nabla_{\mathbf{P}'_{rgb}} \mathcal{L}_{pqgd}$
        \STATE $Balanced \, \, \nabla_{\mathbf{P}'_d} \mathcal{L}_{pqgd} \leftarrow w_d, \nabla_{\mathbf{P}'_d} \mathcal{L}_{pqgd}$
        \STATE $Updated \, \, \mathbf{p}'_{rgb} \leftarrow \mathbf{p}'_{rgb}, Balanced \, \, \nabla_{\mathbf{P}'_d} \mathcal{L}_{pqgd}$
        \STATE $Updated \, \, \mathbf{p}'_d \leftarrow \mathbf{p}'_d, Balanced \, \, \nabla_{\mathbf{p}'_d} \mathcal{L}_{pqgd}$
\ENDFOR
\RETURN $(\mathbf{p}'_{rgb}, \mathbf{p}'_d)$
\end{algorithmic}
\end{algorithm}

Based on sensitivity analysis from the whole image, we conduct reweighting for the gradient from RGB $\nabla_{\mathbf{p}'_{rgb}} \mathcal{L}_{pqgd}$ and depth patches $\nabla_{\mathbf{p}'_d} \mathcal{L}_{pqgd}$ in the human hand area, as shown in Eq. \ref{eq:17} and Eq. \ref{eq:18}. Here $w_{rgb}$ and $w_{d}$ are reweighting parameters based on the sensitivity ratio, which is set to $\rho$ and 1, respectively. Therefore, this operation can let the gradient from the RGB patch approach the gradient from the depth patch within the human hand area, realizing balanced gradient contribution.

\begin{equation}
\label{eq:17}
    \nabla_{\mathbf{p}'_{rgb}}^{b} \mathcal{L}_{pqgd} = w_{rgb} \cdot \nabla_{\mathbf{p}'_{rgb}} \mathcal{L}_{pqgd}
\end{equation}

\begin{equation}
\label{eq:18}
    \nabla_{\mathbf{p}'_d}^{b} \mathcal{L}_{pqgd} = w_d \cdot \nabla_{\mathbf{p}'_d} \mathcal{L}_{pqgd}
\end{equation}

Finally, GLMBS also incorporates the distance-adaptive perturbation bound $\epsilon'(d)$ for the depth, as shown in Eq. \ref{eq:19}. Where $\epsilon$ is a fixed perturbation bound for RGB, the same as in Eq. \ref{eq:9} and Eq. \ref{eq:10}, $d$ denotes one variable depth value, $d_{\max}$ and $d_{\min}$ denote the maximum and minimum depth value in the depth image, $\lambda$ is a hyperparameter. This operation can maintain fixed perturbation bounds for RGB while allowing adaptive bounds for depth based on measured distance, reflecting the different noise characteristics of each modality, and further enhancing the shape adaptation process.

\begin{equation}
\label{eq:19}
\epsilon'(d) = \epsilon \left( 1 + \lambda \frac{d - d_{\min}}{d_{\max} - d_{\min}} \right)
\end{equation}

\subsection{Implement in the Real Robot Grasping Platform}

We follow the same deployment procedure as in QFAAP \cite{c9} to implement our method on a real robot grasping platform for RGBD-based safe grasping. Specifically, after integrating MAQP into the system, the human hand will serve as a suppression source to reduce the predicted quality scores of objects in its vicinity. Meanwhile, the quality scores within the hand region will be set to zero. As a result, the robot will be guided to grasp objects that are away from the human hand and its nearby objects, thereby enhancing HRI safety without emergency stops.

\section{Experiments} \label{sec:ex}
In this section, we first provide a detailed description of the experiment setting, includes setting for MAQP and settings for robot grasping. Then, we validate the effectiveness of MAQP in different models and datasets, followed by the ablation studies to validate each component of MAQP. Finally, we test our method in real grasping with hand dynamic interference.

\subsection{Experimental Settings}
\subsubsection{Setting for MAQP} \label{sec:smaqp}
We perform our experiments on the Cornell Grasp Dataset \cite{c24} and the OCID Grasp Dataset \cite{c25}. The Cornell Grasp Dataset is a single-object RGBD dataset comprising 885 image pairs, whereas the OCID dataset contains cluttered scenes with approximately 1.7k image pairs.

\begin{figure}[!t]
\vspace{0.5\baselineskip}
\centerline{\includegraphics[width=\columnwidth]{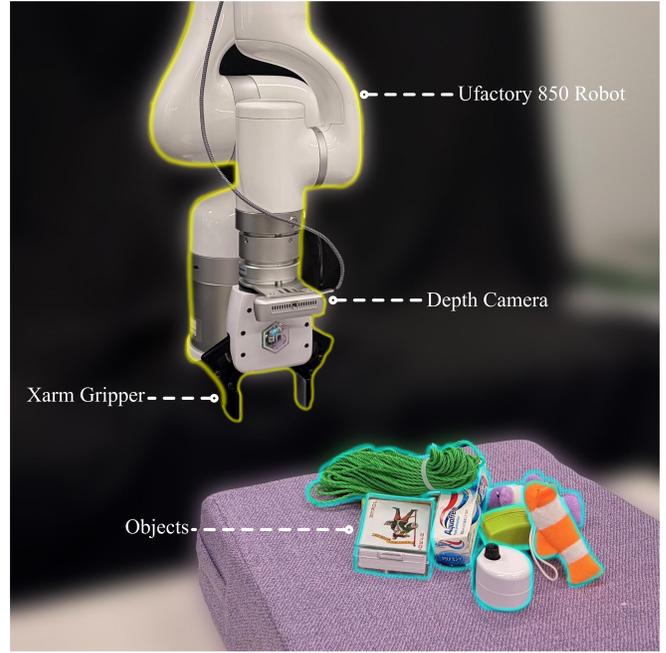}}
\caption{Robot grasping platform: consisting of an Intel RealSense D435 depth camera, a UFactory 850 robot, a UFactory xArm gripper, and the experimental objects}
\label{fig3}
\end{figure}

To facilitate RGBD patch generation, we pretrain several representative DNNs-based grasping networks, including GG-CNN \cite{c26}, GG-CNN2 \cite{c27}, GR-ConvNet \cite{c12}, FCG-Net \cite{c28}, and SE-ResUNet \cite{c29}. Among these models, GR-ConvNet and SE-ResUNet directly support RGBD inputs, FCG-Net is originally designed for RGB images, and the GG-CNN series operates on depth images. For consistency, we extend FCG-Net and the GG-CNN series to support RGBD data by modifying the input channel dimensions. All models are trained on a single NVIDIA RTX 3090Ti GPU with 24GB memory. Following the image-wise split of GR-ConvNet \cite{c12}, the dataset is randomly shuffled, with 90\% used for training and 10\% for testing. During the training stage, the data will be uniformly cropped to 224$\times$224 (GG-CNN series are 300$\times$300), the total number of epochs for training is set to 50, the learning rate $\delta_{model}$ is fixed to 0.001, batch size $B$ is set to 8, and data augmentation is applied.

For the generation of the RGBD patch, we use the same device, system, and training parameters as the grasping model. Differently, we first initialize the RGB patch with a uniform distribution and the depth patch with gaussian distribution of size 224$\times$224 (300$\times$300 for GG-CNN series) as discussed in HDPOS. Next, during each iteration, we apply a random scale to the patch and paste it onto a random position of the training sample that is consistent with \cite{c9}. We also set $\alpha$, $\beta$, and $\gamma$ in $\mathcal{L}_q^p$ and $\mathcal{L}_{aqp}$ to 0.1, 0.1, and 0.5 like \cite{c9}. The initial learning rate $\delta_{aqp}$ is set to 0.03 (decreasing by a factor of ten at the 30th and 40th epochs), the same as in \cite{c9}. Finally, we evaluate the performance of the RGBD patch on the test set using the Q-ACC in \cite{c9}.

\begin{table}[!t]
\centering
\caption{RESULTS OF MAQP ON THE CORNELL GRASP DATASET\label{tab:table1}}
\begin{tabular}{lccc}
\toprule
Models &O-ACC (\%) & Q-ACC (\%) & Runtime ($s$)\\
\midrule
GG-CNN & 89.9 & 70.1 & 0.004 \\
GG-CNN2 & 94.9 & 90.3 & 0.010 \\
GR-ConvNet & 96.2 & 85.9 & 0.016 \\
FCG-Net & 94.9 & 89.5 & 0.047 \\
SE-ResUNet & 97.5 & 97.2 & 0.047 \\
\bottomrule
\end{tabular}
\end{table}

\begin{table}[!t]
\centering
\caption{RESULTS OF MAQP ON THE OCID GRASP DATASET\label{tab:table2}}
\label{tab:ocid_results_2}
\begin{tabular}{lccc}
\toprule
Models &O-ACC (\%) & Q-ACC (\%) & Runtime ($s$)\\
\midrule
GG-CNN & 28.8 & 85.7 & 0.006 \\
GG-CNN2 & 43.5 & 97.6 & 0.012 \\
GR-ConvNet & 58.2 & 87.0 & 0.038 \\
FCG-Net & 59.3 & 92.5 & 0.051 \\
SE-ResUNet & 60.5 & 90.1 & 0.057 \\
\bottomrule
\end{tabular}
\end{table}

For the shape adaptation of the RGBD patch, we set the following parameters that similar to \cite{c9}: the iteration number $N^i$ is set to 1, the learning rate $\delta_{pqgd}$ is fixed at 0.008, $\epsilon$ for RGB patch is set to 8/255, $\epsilon'(d)$ for depth patch is based on variable depth value, and $w_{rgb}$ and $w_d$ are set to $\rho$ and 1 as discussed in GLMBS. In addition, we use the pre-trained model from \cite{c31} for real-time hand segmentation to guide the shape adaptation process of the patch. Finally, since shape adaptation is based on patch generation, we use the same Q-ACC to evaluate the performance of our method.

\subsubsection{Setting for Robot Grasping} \label{sec:egrasp}
Our robot grasping system and part of the experimental objects are illustrated in Fig. \ref{fig3}. For the grasping system, we adopt an eye-in-hand grasping architecture. For the experimental objects, we collected 10 novel objects that are not included in the training dataset. We employ the same Deviation–Return–Deviation Rate (DRD-Rate) in \cite{c9} as evaluation metric, that is, the hand will dynamically approach-away-approach an object with the highest grasp quality score in the camera view, and the distance between the hand and this object remains within 0.5 $cm$, without making physical contact. Finally, we also reproduce the closed-loop control method from \cite{c27} and integrate it into our method, endowing it with the reactive capability to counteract hand dynamic interference.

\subsection{Results on the Different Datasets and Models}

We employ the same experimental setting discussed in Section \ref{sec:smaqp} to validate the performance of MAQP (including both patch generation and shape adaptation process) across different datasets and grasping models. The results are presented in Table \ref{tab:table1} (results in Cornell Grasp Dataset) and Table \ref{tab:table2} (results in OCID Grasp Dataset). In these results, MAQP optimized by most models and datasets achieves a Q-AAC exceeding 85\%, except for those optimized by GG-CNN and the Cornell Grasp Dataset, which attains 70.1\%. The runtime indicates that our method can also achieve real-time speed in predicting a single MAQP. The above analyses indicate that MAQP in different datasets and models is effective.

\begin{table}[!t]
\centering
\caption{The Impact of Modality-Specific Initialization on Q-ACC\label{tab:table3}}
\label{tab:ablation_init}
\begin{tabular}{lccc}
\toprule
Models & \makecell{Q-ACC (\%) \\Fixed} & \makecell{Q-ACC (\%) \\Adaptive} & Runtime ($s$)\\
\midrule
GG-CNN2 & 96.3 & 94.9 & 0.028 \\
GR-ConvNet & 93.4 & 94.4 & 0.015 \\
FCG-Net & 89.6 & 92.4 & 0.019 \\
\bottomrule
\end{tabular}
\end{table}

\begin{table}[!t]
\centering
\caption{The Impact of Gradient Reweighting on Q-ACC\label{tab:table4}}
\begin{tabular}{lccc}
\toprule
Models & \makecell{Q-ACC (\%)\\Unbalanced} & \makecell{Q-ACC (\%) \\Balanced} & $\rho$\\
\midrule
GG-CNN2  & 90.6 & 93.1 & 1.03\\
GR-ConvNet  & 91.5 & 93.0 & 1.02\\
FCG-Net  & 88.4 & 85.3 & 0.99\\
\bottomrule
\end{tabular}
\end{table}

\begin{table}[!t]
\centering %
\caption{The Impact of Adaptive Perturbation Bound on Q-ACC\label{tab:table5}}
\begin{tabular}{lccc}
\toprule
Models & \makecell{Q-ACC (\%) \\Static $\epsilon$} & \makecell{Q-ACC (\%) \\Adaptive $\epsilon$} & Runtime ($s$)\\
\midrule
GG-CNN2 & 87.5 & 90.3 & 0.010 \\
GR-ConvNet & 88.2 & 89.5 & 0.047 \\
FCG-Net & 89.2 & 85.9 & 0.016 \\
\bottomrule
\end{tabular}
\end{table}

\begin{table}[!t]
\centering
\caption{Grasping Results in Different Scenarios\label{tab:table6}}
\label{tab:real_robot_results}
\begin{tabular}{lcccccc}
\toprule
Patches & S1 & S2 & S3 & S4 & S5 & DRD-Rate (\%) \\
\midrule
Original-generated & 3/5 & 4/5 & 4/5 & 5/5 & 5/5 & 84\%\\
Shape-adapted & 4/5 & 5/5 & 5/5 & 4/5 & 5/5 & 92\% \\
\bottomrule
\end{tabular}
\end{table}

\subsection{Ablation Studies} \label{sec:ablation}

We conduct ablation studies in this part to evaluate the effectiveness of each of the components of the proposed MAQP framework. The experimental setting is consistent with Section \ref{sec:smaqp}. First, we test the HDPOS in the patch generation process, and the results are presented in Table \ref{tab:table3}, indicating that most of the Q-ACC is clearly increased after model-specific initialization, except for the Q-ACC in GG-CNN2 that caused by the design nature for depth modality since we just extend it to solve RGBD modality. In addition, the system can also run in real-time speed after adding HDPOS based on the Runtime results. These results demonstrate that the HDPOS is important in initializing the RGBD patch.

Next, we validate the GLMBS in the shape adaptation process of the RGBD patch, which includes two steps: the gradient reweighting and distance-adaptive perturbation. Here, we evaluate them separately. The results for gradient reweighting are shown in Table \ref{tab:table4}. Our method can not only enhance the Q-ACC (except for FCG-Net because of its RGB modality nature, similar to the problem in GG-CNN2 of Table \ref{tab:table3}), but also successfully push the sensitivity ratio $\rho$ close to 1. The results for the adaptive perturbation bound are shown in Table \ref{tab:table5}, which also indicates improved performance after use adapative bound and can run at real-time speed, except for FCG-Net because of its RGB modality nature. All of these results demonstrated the effectiveness of GLMBS in the shape adaptation process of the RGBD patch.

\begin{figure*}[!t]
\vspace{0.5\baselineskip}
\centerline{\includegraphics[width=\textwidth]{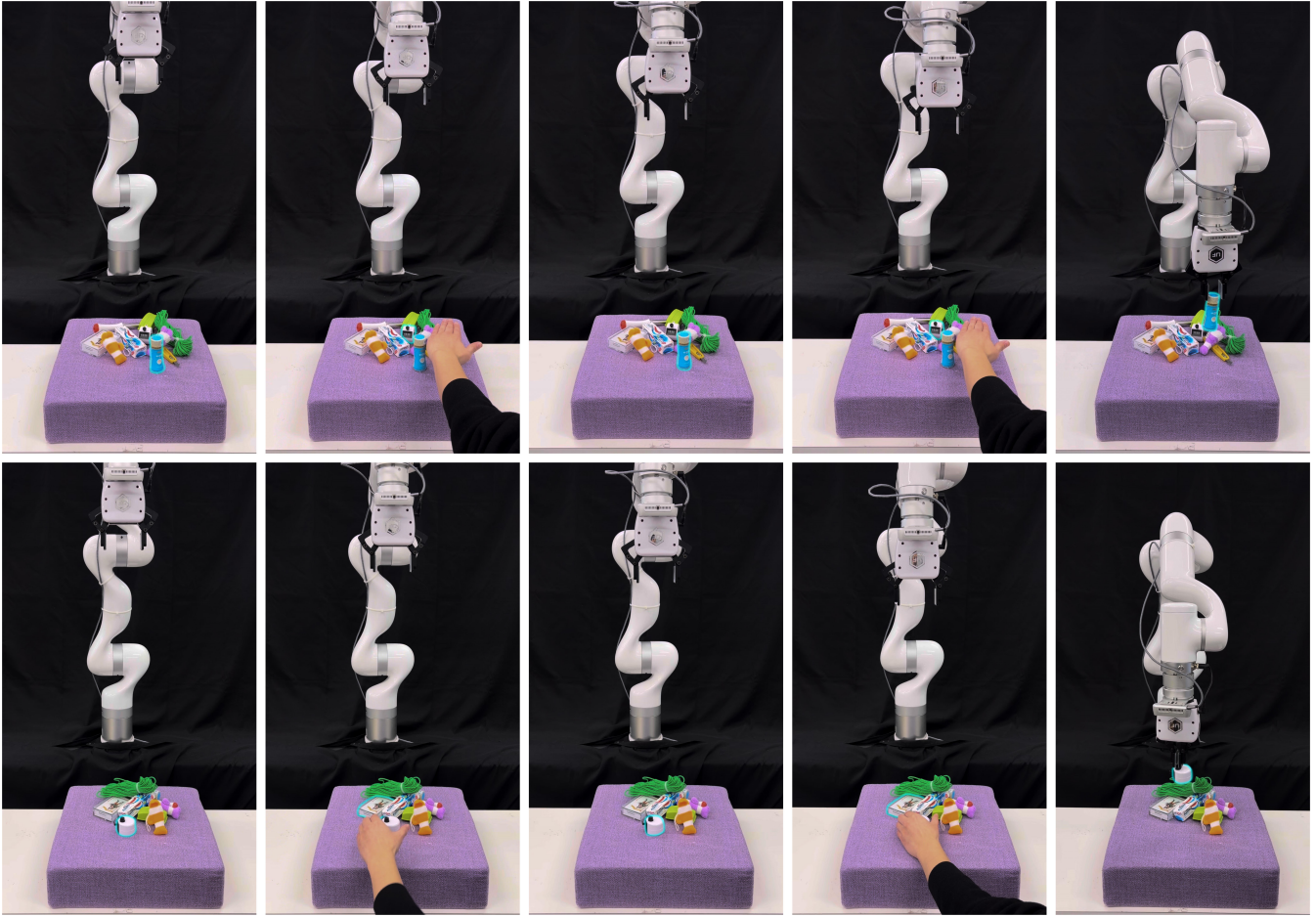}}
\caption{Cases of the DRD process during grasping. Each row represents a single case. From left to right, the images in each row show: the robot’s initial approach to the target object, the first deviation caused by interference, the robot’s re-approach (return) to the target after the human hand moves away, the second deviation following the subsequent interference, and the final grasping after the human hand moves away again. We use the blue border to emphasize the target object that the robot is trying to grasp in each subfigure.}
\label{fig4}
\end{figure*}

\subsection{Real Robot Grasping Experiments}

In this part, we randomly arrange 10 objects in the workspace and construct 5 different grasping scenes using these objects. For each scene, we perform 5 grasping attempts with the original-generated and shape-adapted RGBD patch by incorporating our method. A grasping is considered successful if the robot completes the whole process of Deviation–Return–Deviation. Other experimetal setting is consistent with Section \ref{sec:egrasp}. The results are shown in Table \ref{tab:table6}. The results show that the original-generated and shape-adapted by incorporating our method achieve very high DRD-Rate, 84\% (21/25) and 92\% (23/25), respectively. This demonstrates that our method can effectively integrate with RGBD patches, whether in the stage of original-generated or shape-adapted, and realize safe grasping by responding to the movement of the human hands in the real world. Finally, we show two cases of the DRD process during grasping in Fig. \ref{fig4}, where it can be seen that the robot consistently avoids the human hand and its nearby objects with hand dynamic interference. More clearly and detailed demonstrations are shown in the videos.

\section{Conclusion} \label{sec:con}
In this paper, we present the Multimodal Adversarial Quality Policy (MAQP). Specifically, we employ the Heterogeneous Dual-Patch Optimization Scheme (HDPOS) to mitigate the distribution discrepancy between RGB and depth modalities during RGBD patch generation by adopting modality-specific initialization strategies. Furthermore, we utilize the Gradient-Level Modality Balancing Strategy (GLMBS) to address the optimization imbalance between RGB and depth patches in the shape adaptation process, by reweighting gradient contributions based on sensitivity analysis and applying distance-adaptive perturbation bounds. Extensive experiments demonstrate the effectiveness of our method across different datasets and models, as well as in real-world grasping scenarios.

Future work can be organized into two primary directions. The first direction aims to tackle the limitations discussed in Section \ref{sec:ablation}, further improving the proposed method. The second direction focuses on extending MAQP to handle the multimodal backdoor attack problem introduced in \cite{c32}.

\end{document}